%% file: manuscript.tex
\documentclass[runningheads]{llncs}

\usepackage{graphicx}
\usepackage{amsmath,amssymb}
\usepackage{booktabs}
\usepackage{subcaption}
\usepackage{siunitx}
\usepackage{multirow}
\usepackage{microtype}
\usepackage[hidelinks]{hyperref}
\usepackage{url}
\usepackage{algorithm}
\usepackage{algpseudocode}
\usepackage[section]{placeins}
\usepackage{float} 
\usepackage{xcolor}
\usepackage{kotex}
\usepackage{makecell}

\newcommand{\logit}{\operatorname{logit}}
\newcommand{\sigm}{\sigma}

\title{BEM: Training-Free Background Embedding Memory for False-Positive Suppression in Real-Time Fixed-Background Camera}

\author{
    Junwoo Park\inst{1,2} \and
    Jangho Lee\inst{1}$^{\star}$ \and
    Sunho Lim\inst{3}$^{\star}$
}

\institute{
    Department of Computer Science and Engineering, Incheon National University, Incheon, Republic of Korea \and
    Army Artificial Intelligence Center, Republic of Korea Army, Gyeryong, Republic of Korea \and
    Department of Computer Science, Texas Tech University, Lubbock, TX, USA \\
    \email{\{moon175611, ubuntu\}@inu.ac.kr, sunho.lim@ttu.edu}
}

\begin{document}
\raggedbottom
\maketitle

\begin{abstract}
Pretrained detectors perform well on benchmarks but often suffer performance degradation in real-world deployments due to distribution gaps between training data and target environments.
COCO-like benchmarks emphasize category diversity rather than instance density, causing detectors trained under \emph{per-class sparsity} to struggle in dense, single- or few-class scenes such as surveillance and traffic monitoring.
In fixed-camera environments, the quasi-static background provides a stable, label-free prior that can be exploited at inference to suppress spurious detections.
To address the issue, we propose \emph{Background Embedding Memory (BEM)}, a lightweight, training-free, weight-frozen module that can be attached to pretrained detectors during inference.
BEM estimates clean background embeddings, maintains a prototype memory, and \emph{re-scores detection logits with an inverse-similarity, rank-weighted penalty}, effectively reducing false positives while maintaining recall.
Empirically, background-frame cosine similarity correlates negatively with object count and positively with \emph{Precision-Confidence AUC (P-AUC)}, motivating its use as a training-free control signal.
Across YOLO and RT-DETR families on LLVIP and simulated surveillance streams, BEM consistently reduces false positives while preserving real-time performance. Our code is available at
\url{https://github.com/Leo-Park1214/Background-Embedding-Memory.git}
\end{abstract}

\section{Introduction}\label{sec:1}
\input{1.introduction}

\section{Motivation: Dataset Bias and Background Priors}\label{sec:2}
\input{2.motivation}

\section{Related Work}\label{sec:3}
\input{3.related_work}
\section{Background Embedding Memory (BEM)}\label{sec:4}
\input{4.methodology}

\section{Experiments}\label{sec:5}
\input{5.experiments}

\section{Discussion}\label{sec:6}        
\input{6.discussion}
\vspace{-0.3em}

\section{Conclusion}\label{sec:7}
\input{7.conclusion}
\vspace{-0.3em}

\section*{Acknowledgements}
This work was supported by the IITP(Institute of Information $\&$ Coummunications Technology Planning $\&$ Evaluation)-ICAN(ICT Challenge and Advanced Network of HRD) grant funded by the Korea government(Ministry of Science and ICT)(IITP-2026-RS-2024-00437024)
\vspace{-0.3em}

\FloatBarrier
\bibliographystyle{splncs04}
{\small\bibliography{references}}

\end{document}

%% file: 1.introduction.tex

Modern object detectors such as YOLO and RT-DETR perform well on COCO and VOC~\cite{lin2014coco,everingham2010voc}, but their precision often drops sharply in surveillance or traffic monitoring deployments due to the \textbf{distribution mismatch} between pretraining datasets and deployment domains~\cite{Torralba2011Unbiased,Inoue2018CrossDomainWSOD,Chen2018DomainAdaptiveFRCNN}.
In many surveillance or industrial deployments, privacy and data-governance constraints also make it difficult to collect or annotate sufficient target-domain data for fine-tuning, even when retraining could mitigate distribution mismatch.

Benchmark datasets emphasize \emph{category diversity} rather than \emph{per-class density}, leading to \emph{per-class sparsity}, where each image contains only a few instances of any given category.
In dense, single- or few-class fixed-camera streams (e.g., pedestrians or vehicles), detectors may misinterpret repetitive background structures or shadows as foreground objects, inflating false positives.
Domain-specific retraining can reduce these errors but requires substantial annotation effort and repeated training cycles, and risks overfitting or catastrophic forgetting~\cite{kirkpatrick2017overcoming,aljundi2019gradient}.

In fixed-camera scenarios, the scene provides a quasi-static \emph{background prior} that remains largely underutilized.
This prior can be quantified via the \textbf{embedding similarity} between a frame and a clean background prototype, expressed as $c = E_I^{\top}E_B$.
Empirical analysis shows that this similarity decreases as the number of objects in the scene increases, and higher similarity correlates with improved stability in P-AUC, indicating more consistent precision across confidence thresholds.
These trends highlight that background–frame similarity can act as a training-free control signal for reducing false positives (see Fig.~\ref{fig:llvip_correlations}).

Motivated by these observations, we propose \textbf{Background Embedding Memory (BEM)}, 
a simple and \emph{training-free} inference-time module that can be attached to existing detectors without modifying any weights.
BEM maintains a lightweight background embedding prototype and applies a similarity-based logit adjustment to suppress background-induced false positives. 
The module operates with negligible latency cost and requires no supervision, additional training, or architectural changes.
Experiments on fixed-camera datasets show that BEM consistently reduces false positives while preserving recall and real-time performance.
The key contributions of this paper are summarized in four-fold:
\begin{itemize}
    \item We analyze how per-class sparsity in common benchmarks harms robustness in dense, single-class fixed-camera environments.
    \item We show that background-frame similarity correlates with scene density and confidence-precision stability, justifying its use as a training-free control signal.
    \item We propose BEM, a lightweight inference-time module that re-scores detector outputs using background–background similarity, without retraining or modifying detector parameters.
    \item We demonstrate consistent false-positive suppression on LLVIP across YOLO and RT-DETR families with minimal computational overhead.
\end{itemize}

%% file: 2.motivation.tex

\input{figures_tex/fig_persons_per_image_distributions}

\subsection{Problem Setting and Assumptions}\label{sec:2.1}
False positives often arise from a distribution mismatch between training and deployment data. 
Multi-class benchmark datasets such as COCO~\cite{lin2014coco} and VOC~\cite{everingham2010voc} contain many categories but few instances per class per image, whereas fixed-camera datasets such as LLVIP~\cite{jia2021llvip} focus on one or two categories with dense occurrences and a stable background.
Fig.~\ref{fig:dataset_diffs} compares persons-per-image distributions across datasets, showing that LLVIP exhibits substantially denser per-class occurrences than COCO and VOC.
We leverage this background stability as a corrective prior for detection score calibration.
We make two assumptions: (1) as the number of objects in a frame increases, the foreground area grows, reducing cosine similarity between frame and background embeddings; (2) dense frames with many objects tend to have lower precision than background-dominant frames.
We empirically confirm on LLVIP that background similarity $c \in [-1,1]$ decreases with object count and that P-AUC increases with $c$ (Fig.~\ref{fig:llvip_correlations}), supporting the use of $c$ as a training‑free control signal for inference‑time confidence re‑scoring. 
Further details are provided in Sec.~\ref{sec:5}.

\subsection{Empirical Validation of Background Similarity}\label{sec:2.2}
To assess whether background–frame similarity can act as a meaningful training-free control signal, we analyze its relationship with scene density and confidence–precision stability on LLVIP.
Empirically, we observe that frames with higher background similarity tend to contain fewer objects and exhibit more stable precision across confidence thresholds, as measured by P-AUC.
These trends support the assumptions introduced in Sec.~\ref{sec:2.1} and motivate the use of background similarity for inference-time confidence re-scoring.
Detailed experimental protocols and quantitative analyses are presented in Sec.~\ref{sec:5.3.1}.

\input{figures_tex/fig_relationships_cosine_similarity}

%% file: figures_tex/fig_persons_per_image_distributions.tex

\begin{figure}[!t]
    \centering
    \begin{subfigure}[b]{0.49\linewidth}
    \includegraphics[width=\linewidth]{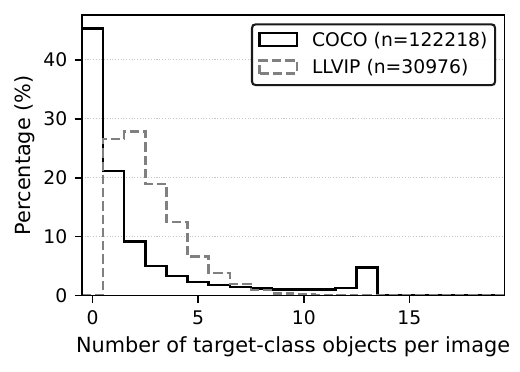}
    \caption{COCO vs LLVIP}
    \label{fig:dataset_coco_llvip}
    \end{subfigure}
    \begin{subfigure}[b]{0.49\linewidth}
    \includegraphics[width=\linewidth]{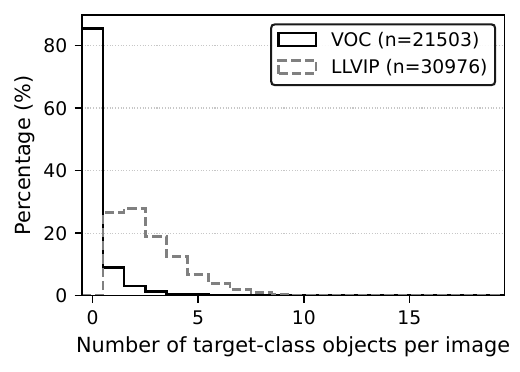}
    \caption{VOC vs LLVIP}
    \label{fig:dataset_voc_llvip}
    \end{subfigure}
    \caption{Persons-per-image distributions comparing COCO/VOC with LLVIP. LLVIP contain substantially denser per-class instances, reflecting real surveillance and traffic scenes~\cite{jia2021llvip}.}
    \label{fig:dataset_diffs}
    \vspace{-1em}
\end{figure}

%% file: figures_tex/fig_relationships_cosine_similarity.tex

\begin{figure}[!t]
  \centering
  \scriptsize
  \begin{subfigure}[b]{0.48\linewidth}
    \includegraphics[height=4.3cm]{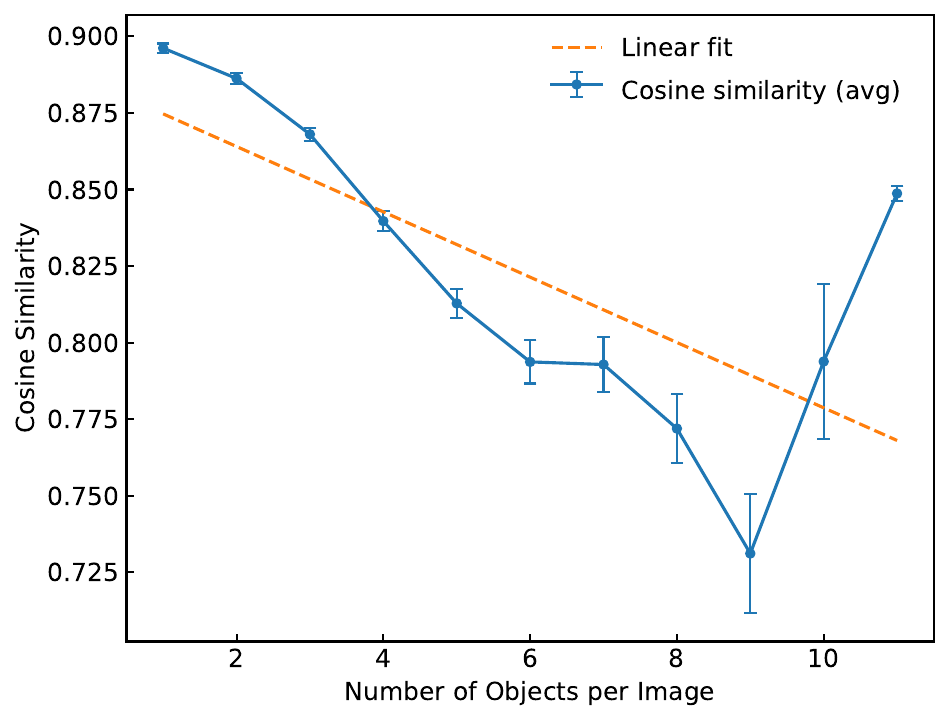}
    \caption{Cosine similarity vs.\ objects per image (negative correlation).}
    \label{fig:llvip_sim_vs_count}
  \end{subfigure}
  \hfill  
  \begin{subfigure}[b]{0.48\linewidth}
    \includegraphics[height=4.3cm]{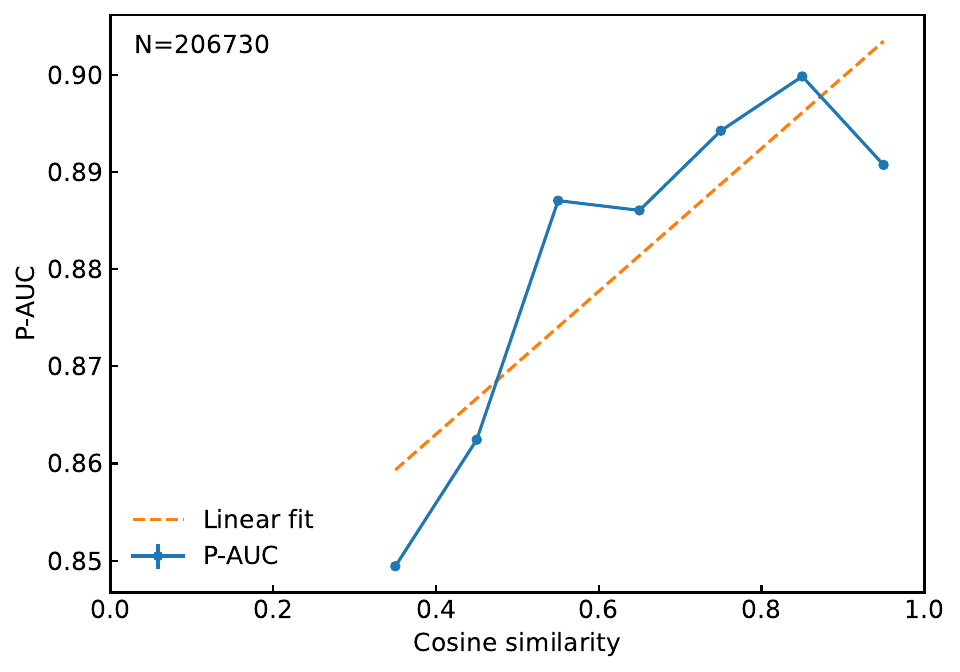}
    \caption{P-AUC vs.\ cosine similarity $c$ (positive association).}
    \label{fig:llvip_pcauc_vs_sim}
  \end{subfigure}
  \caption{Relationship between background--frame cosine similarity $c$ and (a) number of objects, and (b) P-AUC on LLVIP.}
  \label{fig:llvip_correlations}
\end{figure}

%% file: 3.related_work.tex

\subsection{Training-Time FP Suppression and Its Limitations}\label{sec:3.1}
In two-stage detectors, false positives are commonly reduced by refining proposals and filtering low-quality candidates, as exemplified by the Full-Stage Refined Proposal (FRP) algorithm~\cite{guo2025frp}.
In open-vocabulary detectors, BIRDet focuses on handling background samples through background information modeling and partial object suppression~\cite{zeng2025birdet}.
A shared characteristic of these approaches is that they assume access to labeled data in the target (or augmented) domains and often introduce tight coupling between architecture, loss design, and training recipes.
However, such assumptions are frequently violated in practical deployment settings, where privacy regulations and data-governance policies can preclude the collection or long-term retention of raw video data, rendering supervised retraining pipelines largely infeasible despite their potential benefits.
Consequently, most existing FP-suppression techniques operate purely at training time and depend on supervised retraining or explicit architectural modifications, while the paradigm of \emph{training-free, plug-in, background-aware} suppression—especially in fixed-camera deployments—has received comparatively limited attention.

\subsection{Training-Free Memory and Scene-Aware Calibration}\label{sec:3.2}
Memory-augmented methods exploit spatiotemporal cues to facilitate video understanding, typically through external memory modules and learned read–write mechanisms~\cite{xiao2018video,oh2019video,park2020learning}.
Most of these approaches, however, require supervision to learn how the memory is utilized.
PerSense++~\cite{siddiqui2025persense} is closer in spirit to our work in that it investigates a training-free exemplar bank for \emph{segmentation}, but it is not designed for detection score calibration and assumes curated exemplar sets.
Prior memory-based approaches therefore either rely on supervised training or target different tasks (e.g., segmentation), leaving a key limitation unresolved: the absence of an \emph{unsupervised background memory} that can be built online from inference-time data and explicitly used to calibrate \emph{detection} confidence scores in static-camera environments.
In parallel, post-hoc calibration methods learn a mapping from raw logits or uncalibrated scores to well-calibrated probability estimates, with recent work introducing multivariate calibration frameworks and detector-specific strategies~\cite{kuppers2022confidence,popordanoska2024beyond,kuzucu2024calibrationdetectors}.
These methods typically treat the detector as a black box and fit a global calibration function (e.g., via temperature scaling, Platt scaling, or isotonic regression) on labeled validation data, thereby improving confidence reliability without modifying the detector’s learned weights.
However, existing calibrators operate under a \emph{scene-agnostic} assumption and enforce a single global mapping across all frames, neglecting the quasi-static background characteristics of fixed-camera environments and preventing scene-conditioned confidence calibration.
In contrast, our method introduces a lightweight, training-free re-scoring mechanism that leverages background–frame similarity as a scene-aware control signal.

\subsection{CLIP-Based Re-Scoring, Imbalance, and Domain Bias}\label{sec:3.3}
Another line of work leverages large vision–language models such as CLIP to perform semantic similarity re-scoring between image regions and text prompts or concept embeddings~\cite{radford2021clip}.
In open-vocabulary detection, CLIP features are used as auxiliary semantics for adjusting detector logits and improving robustness under category shift.
BIRDet~\cite{zeng2025birdet} is particularly relevant to our use of background information: it introduces Background Information Modeling (BIM) to replace a single fixed background embedding with dynamic, scene-aware background representations derived from CLIP, and uses the resulting similarities to re-weight classifier scores in open-vocabulary detectors, thereby improving the ability to classify oversized or partially visible regions as background.
However, CLIP-based re-scoring and BIM-style background modeling incur significant computational overhead and require extra supervision, making them unsuitable for training-free fixed-camera deployments.
Moreover, they do not exploit per-scene background prototypes derived directly from the detector backbone.
Imbalance and domain shift are additional well-known factors that degrade detection performance~\cite{oksuz2019imbalance,crasto2024cib}.
Typical remedies modify training objectives (e.g., re-weighting or re-sampling) or apply supervised domain adaptation to correct dataset-level statistics, but these strategies primarily address global distribution differences and do not explicitly leverage deployment-specific cues such as fixed-camera backgrounds.
Training-time remedies can improve robustness on average yet overlook the \emph{quasi-static background prior} that is readily available during inference in fixed-camera scenarios, and a deployment-aware, training-free mechanism that directly exploits this prior has been missing.

\paragraph{Positioning of Our Work.}
BEM is a \emph{training-free, weight-frozen} module that:  
(i) builds a background prototype from recent unlabeled frames using the detector backbone, and  
(ii) performs logit re-scoring that is modulated by a simple scene-level statistic, namely the background–frame similarity.
Empirically, we show a connection between background similarity and precision stability (P-AUC), and demonstrate that this backbone-native similarity signal can suppress false positives with negligible computational overhead across YOLO and RT-DETR families on LLVIP~\cite{jia2021llvip}.

%% file: 4.methodology.tex

Fig.~\ref{fig:bem_architecture} provides an overview of the proposed \emph{Background Embedding Memory (BEM)} architecture.
BEM is integrated with a pretrained detector at inference time and operates without any modification to the detector's weights.
The framework comprises three stages: (1) background estimation, (2) construction of a background embedding memory, and (3) similarity-driven re-scoring, all performed with the detector weights kept \emph{frozen}.

\subsection{Background Estimation}\label{sec:4.1}
Given a sequence of recent frames $\{I_t\}_{t=1}^{L}$ and the corresponding binary masks $\{M_t\}_{t=1}^{L}$, where $M_t=0$ indicates detected object regions and $M_t=1$ elsewhere, we estimate a clean background image $B$ as:
\begin{equation}
    B = \frac{\sum_{t=1}^{L} I_t \odot M_t}{\sum_{t=1}^{L} M_t}.
\end{equation}
This formulation performs a masked temporal aggregation that suppresses foreground regions while averaging the remaining background pixels across time.
Such masked averaging is a widely adopted strategy for background modeling in fixed-camera scenarios, as it effectively leverages temporal redundancy to recover static scene content.

Similar background extraction mechanisms have been employed in prior work.
In particular, Zhang and Hoai~\cite{Zhang_2023_CVPR} utilize a related masked temporal aggregation scheme within a self-supervised scene adaptation framework for object detection under static camera settings.
In practice, slow illumination variations and gradual scene changes are accommodated through temporal filtering and periodic background refresh, ensuring robustness over extended deployment.

\input{figures_tex/fig_overview}

\subsection{Background Embedding Memory}\label{sec:4.2}
Let $f(\cdot)$ denote the detector backbone.
We extract globally pooled and $\ell_2$-normalized feature embeddings as:
\begin{equation}
    E_B = \mathrm{norm}(\mathrm{pool}(f(B))), \quad E_I = \mathrm{norm}(\mathrm{pool}(f(I))),
\end{equation}
where $B$ denotes background images used for background feature extraction, and $I$ denotes input images that are excluded from the background extraction process.
Here, $E_B$ represents the background embedding, and $E_I$ represents the embedding of the current input frame.
We maintain a single background prototype $\bar{b}$, instantiated as $E_B$, to form the background embedding memory.
The frame-background similarity is computed via cosine similarity: 
\begin{equation}
    c = E_I^\top E_B.
\end{equation}

\subsection{Similarity-Driven Logit Re-Scoring}\label{sec:4.3}
Given raw detection scores $\{s_i\}$ over $N$ object proposals, we first obtain a calibrated confidence $\tilde{s}_i$ by optionally clipping or temperature-based sharpening to $\{s_i\}$.
We further define $r_i\in\{1,\dots,N\}$ as the rank of proposal $i$ when proposals are sorted in descending order of confidence.
We then apply similarity-aware re-scoring in logit space:
\begin{equation}
z_i' = \logit(\tilde{s}_i) - \frac{\alpha}{\gamma}\,\frac{w_i}{\max(c,\delta)}\, ,\qquad s_i'=\sigm(z_i'),
\end{equation}
where $w_i=\bigl(\tfrac{N-r_i}{N+1}\bigr)$ is a rank-based weighting term that assigns smaller penalties to higher-ranked (higher-confidence) proposals.
The hyperparameters $(\alpha,\gamma)$ jointly control the magnitude and sharpness of the penalty: $\alpha$ sets the overall penalty scale, while $\gamma$ acts as a temperature parameter that normalizes the penalty via $\frac{1}{\gamma}$.
The similarity term $c$ reflects background–frame similarity, and $\delta$ is a small positive constant (e.g., $10^{-6}$) introduced to ensure numerical stability.

Subtracting the similarity-weighted penalty in logit space results in a monotonic suppression of proposals that are less consistent with the estimated background.
This formulation is conceptually related to recent approaches on logit normalization and confidence calibration~\cite{wei2022logitnorm,liu2022margin,kuppers2020calibration}, while remaining training-free and inference-time only.

Intuitively, lower background similarity $c$ induces stronger penalties, leading to more aggressive down-weighting in frames containing many objects or undergoing significant background changes.
Conversely, frames with high background similarity are minimally affected.
This frame-level control is further modulated by the proposal-wise rank weighting $w_i$, which prevents over-penalization of the most confident detections.
Here, the hyperparameter $\delta$ serves as a numerical floor to avoid division by zero.

%% file: figures_tex/fig_overview.tex

\begin{figure}[!t]
  \centering
  \includegraphics[width=\linewidth]{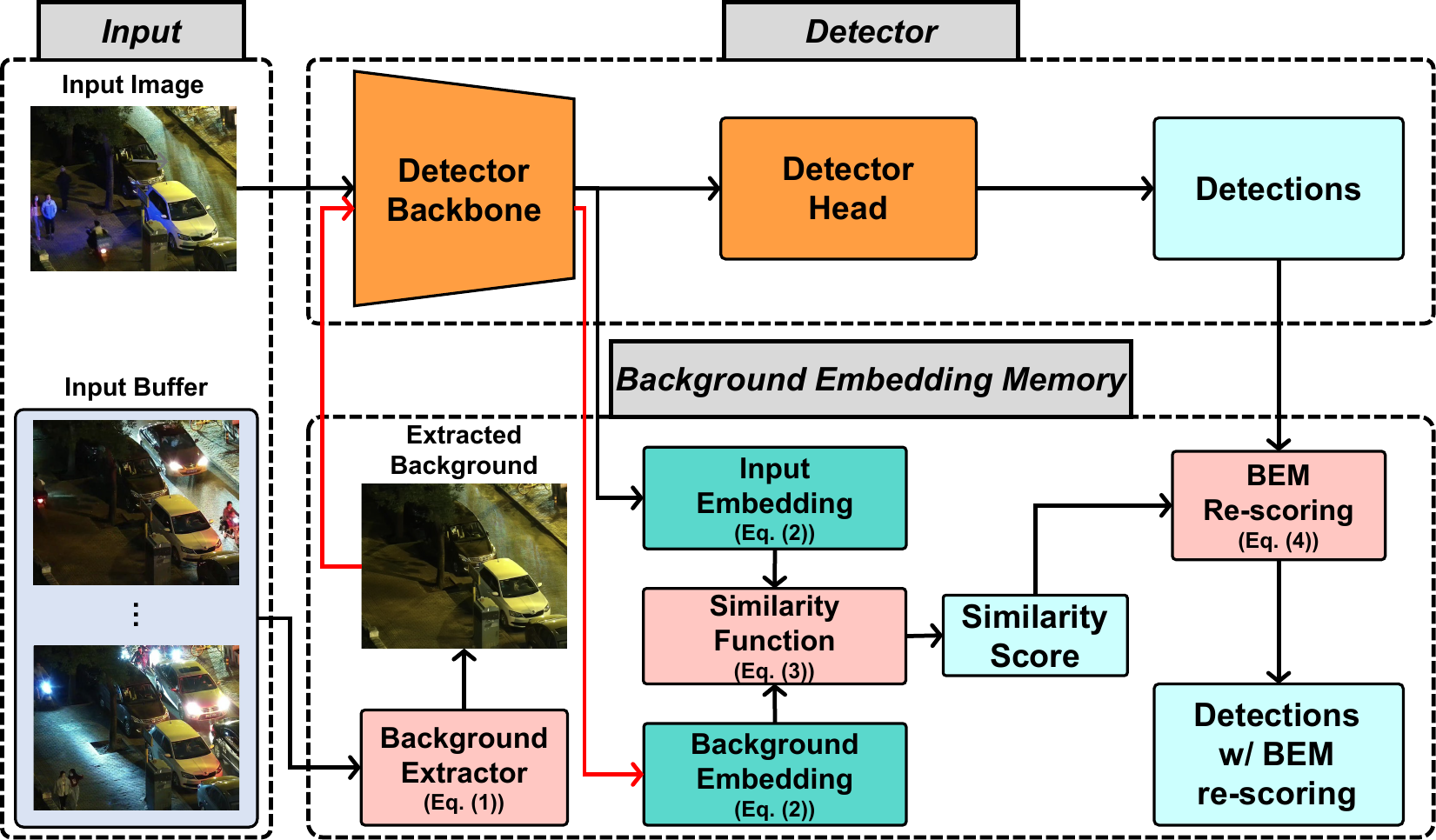}
  \caption{ Overview of the Background Embedding Memory (BEM) pipeline. During inference, BEM attaches to a \emph{frozen} detector without retraining. The module first extracts a background embedding from fixed-scene frames via the shared backbone, computes cosine similarity with the current input embedding, and re-scores the detector’s output logits to suppress background-induced false positives.
  }
  \label{fig:bem_architecture}
\end{figure}

%% file: 5.experiments.tex

\subsection{Dataset and Experimental Setup}\label{sec:5.1}
We evaluate the proposed BEM on the LLVIP dataset~\cite{jia2021llvip}, which represents a fixed-camera surveillance scenario with dense object instances and limited category diversity.
The LLVIP dataset contains $16,836$ paired visible and infrared images captured from $26$ street locations.
Visible images have a resolution of $1920 \times 1080$, while infrared images are $1280 \times 720$,
and each pair is precisely aligned in both time and space.
All image pairs include at least one pedestrian and are annotated with a single pedestrian category.

To assess the generality of BEM across detector architectures, we consider a diverse set of pretrained object detectors, including YOLO v11m~\cite{yolov11_ultralytics_2024}, YOLO v8s~\cite{yolov8_ultralytics_2023}, YOLO-World v2(s, l)~\cite{yoloworld2024}, and RT-DETR~\cite{rtdetr2024}.
All YOLO-family models are initialized using the official pretrained weights released by Ultralytics~\footnote{https://docs.ultralytics.com/}.
For each detector, we evaluate two training variants:
(i) \textbf{COCO-pretrained models}, which directly use the Ultralytics-released checkpoints trained on COCO, and (ii) \textbf{VOC-finetuned models}, obtained by fine-tuning the COCO-pretrained weights on the PASCAL VOC dataset (COCO$\rightarrow$VOC) following the default Ultralytics training recipe with SGD.
This results in a total of six detector variants, all of which are evaluated on LLVIP.

Importantly, all detector weights remain frozen during evaluation.
BEM is applied as an external inference-time module, without any additional training or parameter updates to the underlying detectors.
This setup ensures that any observed performance differences can be attributed solely to the proposed inference-time mechanism rather than retraining effects.
The code will be released available upon publication.

\subsection{Evaluation Metrics}\label{sec:5.2}
We report three evaluation metrics: \textbf{mAP@0.50} to measure standard detection accuracy, \textbf{Precision-Confidence AUC (P-AUC)} to assess confidence-precision stability under confidence thresholding, and \textbf{latency} to evaluate real-time deployment efficiency.
Key hyperparameters of BEM, including the inverse-similarity penalty scale $\alpha$, temperature $\tau$, background window size $L$, and memory size $K$ are summarized in Table~\ref{tab:hyperparams}.
Additional details on the selection of the background window size $L$ are provided in Sec.~\ref{sec:5.3.4}.

\textbf{mAP@0.50.} 
Detection performance is evaluated using the standard COCO-style mean Average Precision (mAP) at an IoU threshold of 0.50.

\textbf{Precision-Confidence AUC (P-AUC).}
To quantify the stability of precision with respect to confidence thresholding, we integrate precision over decreasing confidence thresholds.
Let $\tau \in [0,1]$ denote a confidence threshold; precision $P(\tau)$ is computed using detections with scores greater than or equal to $\tau$.
We define P-AUC as follows:
\begin{equation}
\mathrm{P\text{-}AUC}
\;=\;
\int_{0}^{1} P(\tau)\, d\tau
\;\approx\;
\sum_{j=1}^{J-1} P(\tau_j)\,(\tau_j - \tau_{j+1}),
\end{equation}
where $\{\tau_j\}$ denotes a uniform or score-quantile grid.
Unlike PR-AUC (i.e., AP), which integrates precision over \emph{recall}, P-AUC integrates precision over \emph{confidence}, and thus directly reflects how robust a detector’s precision is to variations in the confidence threshold.

\emph{Interpretation.}
A higher P-AUC indicates that precision remains high across a broad range of confidence thresholds, implying that the detector’s confidence scores are well aligned with correctness.
We further connect P-AUC to background similarity; see Fig.~\ref{fig:llvip_correlations} and the analysis in Sec.~\ref{sec:5.3} for empirical evidence.

\subsection{Main Results}\label{sec:5.3}
Table~\ref{tab:pcauc_results} reports the quantitative comparison of mAP@0.50 and P-AUC between the proposed BEM and the baseline detectors across all evaluated architectures, including YOLO v11~\cite{yolov11_ultralytics_2024}, YOLO v8~\cite{yolov8_ultralytics_2023}, YOLO-World~\cite{yoloworld2024}, and RT-DETR~\cite{rtdetr2024}.
This comprehensive evaluation allows us to assess the consistency of BEM across both detector families and data distributions.

Overall, BEM yields consistent but modest improvements in both mAP@0.50 and P-AUC across all settings.
While the gains in mAP@0.50 are relatively small, they are systematically positive, indicating that the proposed similarity-aware re-scoring does not degrade standard detection accuracy.
At the same time, the observed improvements in P-AUC suggest that BEM effectively stabilizes precision across confidence thresholds, reflecting a reduction in false positives without sacrificing recall.

As further illustrated in Fig.~\ref{fig:llvip_delta_pcauc}, the improvements in P-AUC are not uniformly distributed across all frames.
Instead, they are concentrated in low-similarity, object-dense frames, where the quasi-static background prior is most informative.
This observation is consistent with our underlying assumption that background–frame similarity provides a reliable control signal for suppressing spurious detections in fixed-camera scenarios.

\input{tables_tex/table_main_results}

\input{tables_tex/table_latency_comparison}

\subsubsection{Assumption Validation}\label{sec:5.3.1}
To support the two assumptions stated in Sec.~\ref{sec:2.1}, we analyze how background-frame similarity relates to (i) the number of objects in a scene and (ii) the stability of precision under confidence thresholding.
For each LLVIP frame $I$, we extract a frozen backbone embedding $E_I = \mathrm{norm}(\mathrm{pool}(f(I)))$, using the same detector backbone as in the main experiments.
The background embedding $E_B$ is obtained by the temporal masked aggregation described in Sec.~\ref{sec:4.1}, with the embedding period fixed at the optimal value ($L{=}25$) determined by our background window analysis; this choice ensures that the similarity signal is consistent with the background prior used throughout our evaluation.



\subsubsection{Where Does BEM Improve Precision–Confidence Stability?}\label{sec:5.3.2}
In addition to these raw correlations, we analyze \emph{where} BEM yields the largest improvements in precision-confidence stability. 
For each cosine-similarity bin, we compute the change in P-AUC when enabling BEM (i.e., $\mathrm{P-AUC}_{BEM}$ - $\mathrm{P-AUC}_{Base}$) on LLVIP using YOLO\,v11m (COCO) with the optimal penalty scale~$\alpha$ selected on a held-out validation split.
As shown in Fig.~\ref{fig:llvip_delta_pcauc}, most of the gains arise in low-similarity, object-dense frames-precisely the regime where background-induced false positives are most problematic.

\subsubsection{Selecting the Background Window Size $L$}\label{sec:5.3.3}
We explain how to determine the background window size $L$ used for estimating the clean background $B$.
We empirically swept $L \in \{5,\,10,\,15,\,20,\,25,\,30\}$ on multiple fixed-camera sequences and selected the value that minimized a background-quality score.

\paragraph{Setup.}
For each sliding window of $L$ consecutive frames, we computed a masked temporal average to obtain the background estimate $B$, where foreground regions (identified by detector bounding boxes) were excluded before averaging.  
Given a current image $I$ and background $B$, we computed the residual: $R = \lvert I - B \rvert$, using channel-wise means.

Background quality is evaluated on non-object pixels only using two complementary metrics.
The first metric is the \textbf{mean absolute error (MAE)}, which captures the overall magnitude of residual background noise.
The MAE is calculated by averaging the residual $R$.
The second metric is the \textbf{ghost rate}, defined as the proportion of non-object pixels
whose residual exceeds a fixed threshold ($\tau = 30/255$), reflecting the presence
of ghosting artifacts caused by incomplete foreground suppression or background contamination.
The ghost rate is defined as:
\begin{equation}
  \mathrm{ghost\_rate} = \frac{1}{N_{\text{bg}}}  \sum_{p \in \Omega_{\text{bg}}} \mathbf{1}\bigl[\,R_p > \tau\,\bigr],
\end{equation}
where $\mathbf{1}[\cdot]$ is the indicator function.
These two metrics are combined into a single background-quality score by summing the ghost rate and the MAE with equal weighting.
For each candidate window size $L$, the score is averaged across all sliding windows and all evaluated sequences, and the value of $L$ that yields the lowest mean score is selected as the optimal background window size.

\paragraph{Data and detector.}
We evaluated both infrared and visible streams captured from multiple fixed stations and used a 
single person detector to supply bounding boxes for masking.  
Masked averaging prevented foreground leakage into the background estimate.

\paragraph{Result.}
Across the sweep $L \in \{5, 10, 15, 20, 25, 30\}$, we found that $L=25$ consistently minimized the average score.
This value strikes a balance between temporal robustness (reduced noise and ghosting artifacts) and adaptability to mild illumination changes.
Accordingly, we adopt $L = 25$ as the default background window size for all experiments, unless otherwise specified (see also \hyperref[sec:hyperparams]{Hyperparameters}).

\subsubsection{Hyperparameters for BEM}\label{sec:5.3.4}
This subsection consolidates all hyperparameters used in BEM and the rationale behind their configuration.
The similarity-penalty scale $\alpha$ is tuned via a lightweight grid search, while the temperature $\gamma$ is fixed per detector family to maintain a stable inverse-similarity–based penalty.
The background window size $L$ determines how many recent frames contribute to the masked temporal averaging used to obtain the clean background estimate $B$.
As described in Sec.~\ref{sec:5.3.3}, we select $L=25$ based on an empirical sweep over $L \in \{5,10,15,20,25,30\}$.
Unless stated otherwise, the value of $L$ is used throughout all experiments.
The memory size $K$ controls how many background embeddings are aggregated when forming the prototype.
In our default configuration, we tie $K$ to the background window and set $K=L$, updating the prototype once every $L$ frames.
This makes the prototype refresh period equal to the background-estimation period, simplifying memory behavior without introducing additional hyperparameters.
We also ablate the rank-weighting term $w_i$ to assess its role in false-positive suppression.
As shown in Fig.~\ref{fig:alpha_gamma_ablation}, when the temperature is fixed to $\gamma=0.001$, detection performance becomes highly sensitive to the penalty scale $\alpha$ for both YOLOv11m and YOLOv8l-Worldv2, whereas for larger temperature values, performance remains largely stable across different choices of $\alpha$.

\input{figures_tex/fig_ablation_gamma}

\input{tables_tex/table_hyperparameters}

%% file: tables_tex/table_main_results.tex

\begin{table}[!t]
    \footnotesize
    \caption{ Quantitative comparison of mAP@0.50 and P-AUC across six detectors on the LLVIP dataset.
    \textbf{Base} denotes the baseline detector without BEM, while \textbf{BEM} applies the proposed similarity-driven re-scoring.
    }
    \centering
    \setlength{\tabcolsep}{6pt}
    \begin{tabular}{lcccccc}
    \toprule
    & \multicolumn{2}{c}{P-AUC} 
    & \multicolumn{2}{c}{mAP@0.5} \\
    
    \cmidrule(lr){2-3} \cmidrule(lr){4-5}
    
    Model (Variant) & Base & BEM & Base & BEM \\
    
    \midrule
    
    YOLO\,v11m (COCO)            & 89.82 & \textbf{92.87} \scriptsize{(±0.034)} & 80.49 & \textbf{80.99} \scriptsize{(±0.001)} \\
    YOLO\,v8s (COCO)             & 88.44 & \textbf{91.63} \scriptsize{(±0.017)} & 75.34 & \textbf{75.90} \scriptsize{(±0.028)} \\
    RT-DETR-L (COCO)             & 77.60 & \textbf{82.85} \scriptsize{(±0.030)} & 79.26 & \textbf{79.59} \scriptsize{(±0.022)} \\

    \hline
    
    YOLO\,v11m (COCO$\rightarrow$VOC) & 93.39 & \textbf{94.24} \scriptsize{(±0.004)} & 68.71 & \textbf{69.51} \scriptsize{(±0.012)} \\
    YOLO\,v8s (COCO$\rightarrow$VOC)  & 92.67 & \textbf{93.51} \scriptsize{(±0.013)} & 66.17& \textbf{66.88} \scriptsize{(±0.021)} \\
    RT-DETR-L (COCO$\rightarrow$VOC)  & 78.44 & \textbf{84.19} \scriptsize{(±0.027)} & 66.19 & \textbf{66.58} \scriptsize{(±0.027)} \\

    \hline
    
    YOLO\,v8s-Worldv2                 & 81.78 & \textbf{81.88} \scriptsize{(±0.011)} & 90.23 & \textbf{91.36} \scriptsize{(±0.001)} \\
    YOLO\,v8l-Worldv2                 & 86.22 & \textbf{86.27} \scriptsize{(±0.005)} & 91.20 & \textbf{92.36} \scriptsize{(±0.003)} \\
    
    \bottomrule
    \end{tabular}
    \label{tab:pcauc_results}
\end{table}
\vspace{2em}

%% file: tables_tex/table_latency_comparison.tex

\begin{figure}[!t]
    \centering
    \begin{minipage}[t]{0.48\linewidth}
    \vspace{-40.0mm}
      \centering
      {\fontsize{8.0}{6.0}\selectfont
      \setlength{\tabcolsep}{4pt}
      \captionof{table}{ Latency comparison between baseline detectors and BEM-equipped variants.
      \textbf{Base} denotes the original detector without BEM, while \textbf{BEM} applies the proposed inference-time re-scoring.
      }
        \begin{tabular}{@{\hskip 0.015in}l@{\hskip 0.015in}
                        @{\hskip 0.015in}c@{\hskip 0.015in}}
        \toprule
        Model & \makecell{ Latency \\ (ms/frame) } \\
        \midrule
        YOLOv11m (Base)         & 370.15 (±1.22) \\
        YOLOv11m (BEM)              & 415.02 (±3.81) \\
        
        \midrule
        YOLOv8s (Base)          & 318.49 (±1.97) \\
        YOLOv8s (BEM)               & 368.26 (±5.93) \\
        
        \midrule
        RT-DETR-l (Base)        & 30.87 (±0.14) \\
        RT-DETR-l (BEM)             & 54.44 (±0.34) \\
        
        \midrule
        YOLOv8s-Worldv2 (Base) & 23.52 (±0.10) \\
        YOLOv8s-Worldv2 (BEM)      & 41.67 (±0.12) \\
        \midrule
        YOLOv8l-Worldv2 (Base) & 25.51 (±0.08) \\
        YOLOv8l-Worldv2 (BEM)      & 44.44 (±0.08) \\
        \bottomrule
        \end{tabular}
      \label{tab:deploy_results}
      }
    \end{minipage}
    \hfill
    \begin{minipage}[!t]{0.48\linewidth}
      \centering
      \includegraphics[width=\linewidth]{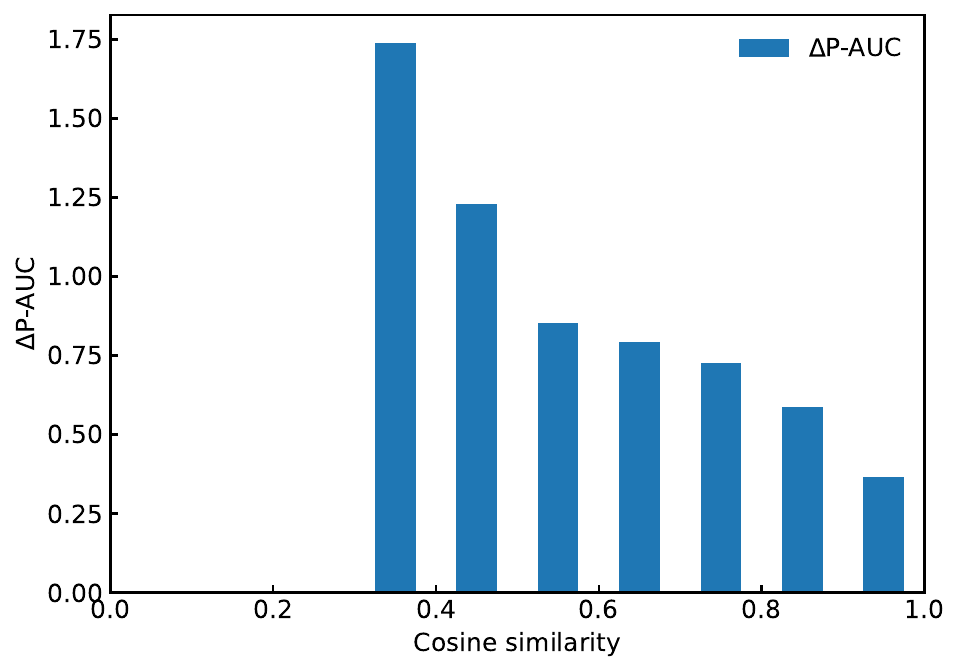}
      \caption{ Increase in P-AUC as a function of background-frame cosine similarity.
      Frames are grouped into similarity bins, and the P-AUC improvement is averaged within each bin.
      Lower similarity corresponds to object-dense frames, where BEM yields larger precision gains. 
      }      
      \label{fig:llvip_delta_pcauc}
    \end{minipage}
\end{figure}

%% file: figures_tex/fig_ablation_gamma.tex

\begin{figure}[!t]
    \small
  \centering
  \begin{subfigure}[b]{0.48\linewidth}
    \includegraphics[width=\linewidth]{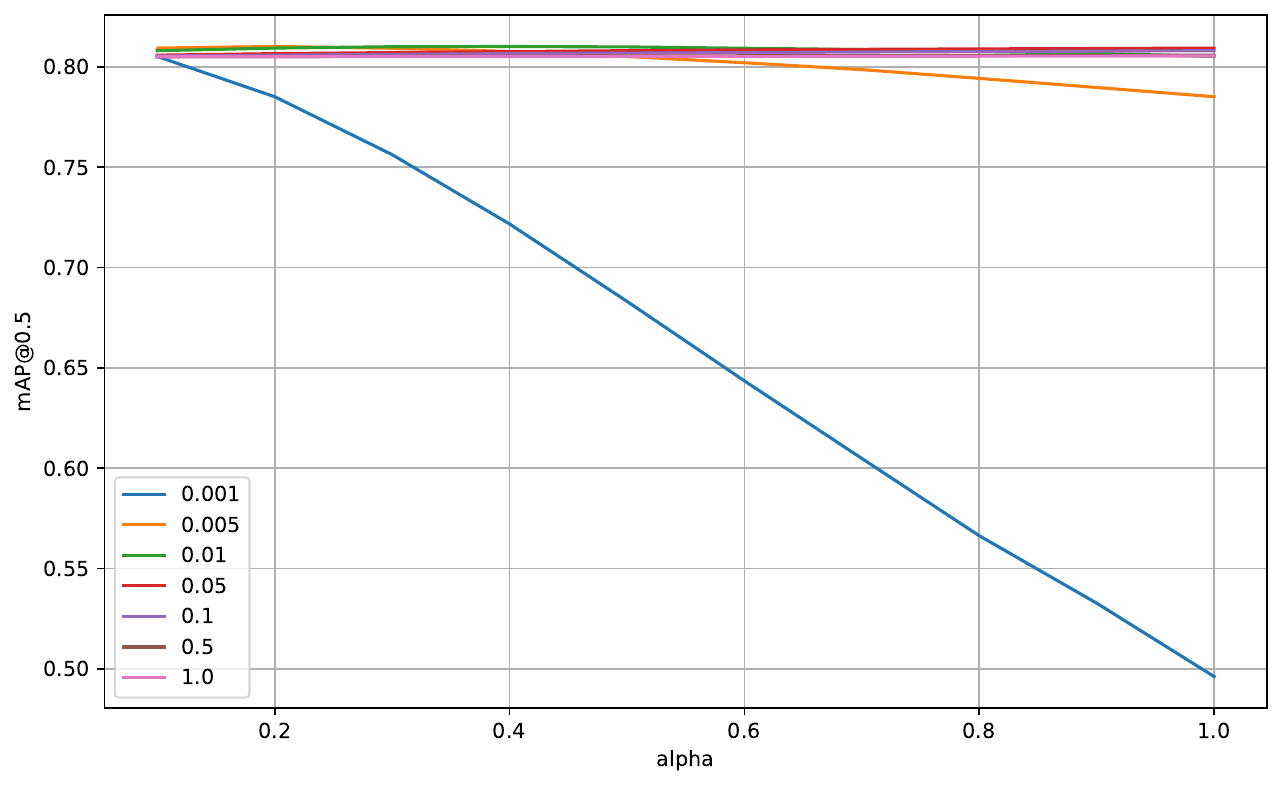}
    \caption{YOLO\,v11m (COCO).}
    \label{fig:alpha_gamma_yolo11m}
  \end{subfigure}\hfill
  \begin{subfigure}[b]{0.48\linewidth}
    \includegraphics[width=\linewidth]{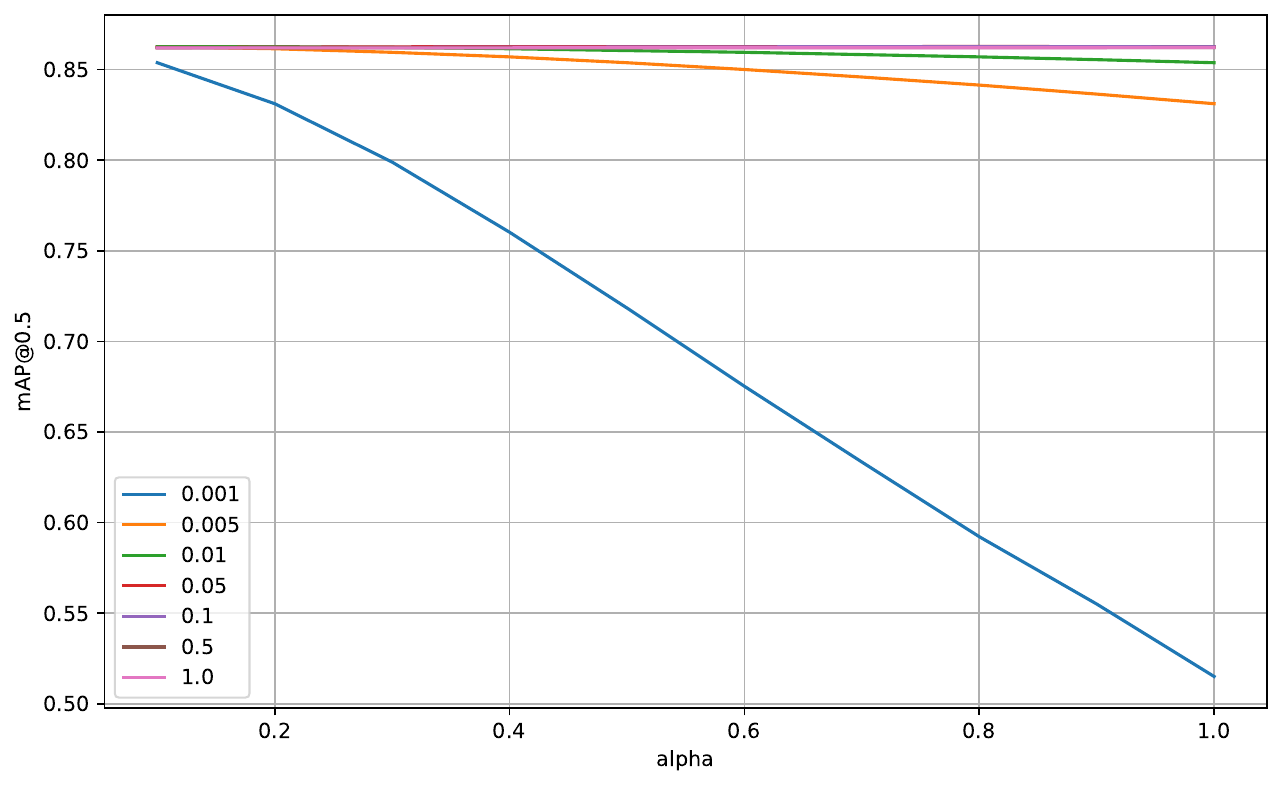}
    \caption{YOLO\,v8l-Worldv2.}
    \label{fig:alpha_gamma_yoloworld}
  \end{subfigure}
  \caption{
  Ablation study of the BEM penalty scale $\alpha$ and temperature $\gamma$ across different detector backbones.
  Each curve reports mAP@0.50 as a function of $\alpha$ for several fixed values of $\gamma$.
  }
  \label{fig:alpha_gamma_ablation}
  \vspace{-1em}
\end{figure}

%% file: tables_tex/table_hyperparameters.tex

\begin{table}[!t]
    \footnotesize
    \centering
    \caption{ Summary of BEM hyperparameters.
    $L$ denotes the number of recent frames used for temporal background estimation, while $K$ denotes the number of background embeddings aggregated to form a single prototype.
    In our experiments, we set $K=L$. }
    \label{tab:hyperparams}
    \begin{tabular}{lcll}
    \toprule
    \textbf{Symbol} & \textbf{Range} & \textbf{Role} & \textbf{Description} \\
    \midrule
    $\alpha$ & [0.0,1.0] & Penalty scale & Controls suppression strength in logit space \\
    $\gamma$ & (0.0,1.0] & Temperature & Normalizes penalty magnitude (sharpness) \\
    $L$ & - & Temporal window & Number of frames for background averaging \\
    $K$ & - & Prototype count & Number of frames aggregated for one prototype \\
    \bottomrule
    \end{tabular}
\end{table}

%% file: 6.discussion.tex

\subsubsection{Deployment Interpretation}\label{sec:6.1}
We consider a common deployment scenario in which pretrained object detectors are applied
to fixed-camera environments in a strict zero-shot manner, without any fine-tuning or supervision
on the target domain, often due to practical constraints such as privacy regulations, data ownership, or the infeasibility of collecting and annotating target-domain data after deployment.
In such settings, detectors that perform well on standard benchmarks often exhibit a noticeable increase in false positives.
Our key observation is that a substantial portion of these false positives arises not from
insufficient model capacity, but from dataset bias induced by benchmark training.

\input{figures_tex/fig_examples}

\subsubsection{Why Similarity Works}\label{sec:6.2}
In particular, detectors trained on datasets with strong per-class sparsity are prone to
\emph{over-confident} predictions on repetitive background patterns when deployed in dense, single- or few-class fixed-camera scenes.
From this perspective, background-frame similarity is not merely a descriptive statistic, but an actionable signal that reflects scene difficulty and susceptibility to background-induced errors. 
BEM addresses this issue by introducing a deployment-aware, training-free re-scoring mechanism that leverages background-frame embedding similarity computed from a frozen backbone.
This similarity serves as a scene-level control signal, allowing confidence scores to be modulated according to how strongly a frame deviates from its background prototype.
To support this interpretation, we examined the relationship between background similarity, scene density, and precision-confidence stability, followed by ablation analyses that justify the choice of the background window size and penalty hyperparameters.
We further analyze where BEM yields the largest gains and under which conditions its assumptions begin to break down.


%% file: figures_tex/fig_examples.tex

\begin{figure}[!t]
  \centering
  \includegraphics[width=\linewidth]{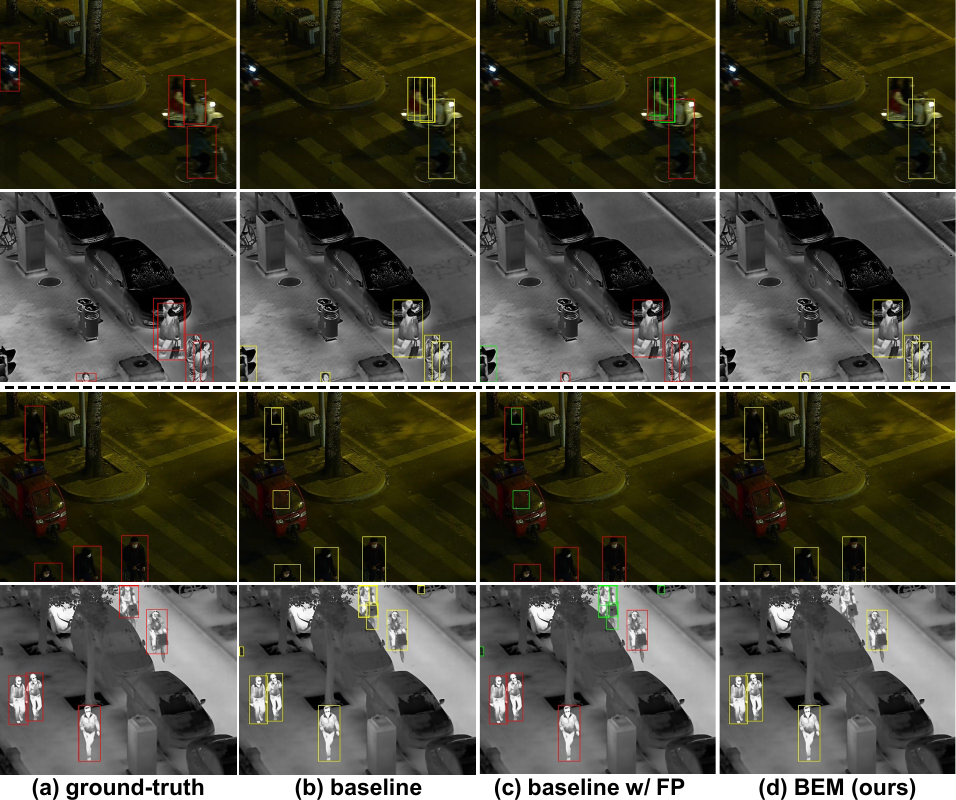}
  \caption{ Qualitative comparison of detection results. The top two rows show results from YOLO-based detectors, while the bottom two rows correspond to DETR-based detectors.
  (a) Ground-truth bounding boxes, (b) Detection results of the baseline detector, (c) Baseline detections with false positives highlighted with green color, and (d) Detection results with the proposed BEM applied.
  }
  \label{fig:bem_examples}
\end{figure}

%% file: 7.conclusion.tex

In this study, we introduced \emph{Background Embedding Memory} (BEM), a simple, training-free inference-time module for reducing false positives in fixed-camera deployment scenarios.
By leveraging background--frame embedding similarity from \emph{frozen} detectors,
BEM suppresses background-induced errors without additional supervision or retraining.
Experiments in controlled fixed-camera settings show that BEM consistently reduces false positives while preserving recall and real-time performance across multiple detectors.
These results demonstrate the effectiveness of background-aware confidence modulation for improving detector reliability under strict zero-shot constraints.
While BEM is primarily designed for relatively stable scenes, it provides a practical foundation for deployment-time reliability enhancement.
Future extensions include more localized similarity measures and adaptive memory updates to handle long-term background drift and abrupt illumination changes.